\documentclass[journal,comsoc]{IEEEtran}
\usepackage[T1]{fontenc}
\usepackage[cmintegrals]{newtxmath}
\usepackage{comment}
\usepackage{algorithmic}
\usepackage{algorithm}
\usepackage{graphicx}
\usepackage{amsmath}
\usepackage{booktabs}
\usepackage{cite}
\usepackage[caption=false,font=footnotesize]{subfig}
\begin{document}
\bstctlcite{IEEEexample:BSTcontrol}

\title{Tram-FL: Routing-based Model Training for Decentralized Federated Learning}

\author{Kota~Maejima,~
        Takayuki~Nishio,~
        Asato~Yamazaki,~
        and~Yuko~Hara-Azumi~

\thanks{This work was supported by JST, PRESTO Grant Number JPMJPR2035, Japan.}

\thanks{K. Maejima, T. Nishio, A. Yamazaki, Y. Hara are with the School of Engineering, Tokyo Institute of Technology, Tokyo 152-8550, Japan (e-mail: maejima.k.aa@m.titech.ac.jp; nishio@ict.e.titech.ac.jp; yamazaki.a.ah@m.titech.ac.jp; hara@cad.ict.e.titech.ac.jp).}% <-this % stops a space
}

% The paper headers
\markboth{}%
{Shell \MakeLowercase{\textit{et al.}}: Bare Demo of IEEEtran.cls for IEEE Journals}

% make the title area
\maketitle

\begin{comment}
データプライバシーとセキュリティに関する新たな懸念は、ノードが自身のオリジナルデータの代わりに、ローカルに訓練されたモデルのみを同期することを可能にする、連合学習の提案を動機付けました。しかし、ノード間のデータが統計的に不均一であるような状況では、モデルの勾配が顕著に異なるため、しばしば局所最小値に引っかかる。
この問題を解決するために昨今様々な手法が提案されている。しかし、データの偏りが著しく大きいnon-IID設定においては、精度の高いグローバルモデルを得ることは困難である。
またネットワーク全体で頻繁に通信が行われるため通信コストが大きくなる傾向にある。本論文では、この問題に対処するために、traveling model training for decentralized federated learning (Tram-FL)と、このアルゴリズムにおけるモデルの巡回経路の決定法としてdynamic model routing algorithmを提案する。Tram-FLは各ノードの持つデータの分布をもとに巡回経路を決定し, １つのモデルをノード間で巡回させながら順番に訓練する。提案するdynamic model routing algorithmはモデル訓練で使用したデータの分布バイアスを最小化する方法で次のノードを選択する。実験の結果、提案アルゴリズムは、ノード上に統計的に非常に異質なデータセットを配置しても、グローバルな認識モデルに少ない通信コストで収束することが示された。

分散型連合学習では、頻繁に通信を行うため通信コストが増大することが多く、ノード間でデータ分布が統計的に不均一な場合、正確な大域モデルを得ることができない。我々は、事前にモデルの巡回経路を決定し、ノード間を巡回しながらモデルを順次学習する巡回モデル学習アルゴリズム(Tram-FL)を提案する。また、学習に用いたラベル分布の偏りを最小化して次のノードを選択する動的モデルルーティングアルゴリズムを提案する 。提案アルゴリズムは、高度に異質なデータセットがノード間に分散している場合でも、最小限の通信コストで正確なグローバル認識モデルを得ることができる。

\end{comment}

\begin{abstract}
In decentralized federated learning (DFL), substantial traffic from frequent inter-node communication and non-independent and identically distributed (non-IID) data challenges high-accuracy model acquisition. We propose Tram-FL, a novel DFL method, which progressively refines a global model by transferring it sequentially amongst nodes, rather than by exchanging and aggregating local models. We also introduce a dynamic model routing algorithm for optimal route selection, aimed at enhancing model precision with minimal forwarding. Our experiments using MNIST, CIFAR-10, and IMDb datasets demonstrate that Tram-FL with the proposed routing delivers high model accuracy under non-IID conditions, outperforming baselines while reducing communication costs.
\end{abstract}

%Keyword

\IEEEpeerreviewmaketitle

\section{Introduction}
\IEEEPARstart{T}{HE} advancements in deep learning algorithms, driven by abundant data, have led to highly accurate models in natural language processing and computer vision. In scenarios where data constraints inhibit the development of high-performance predictive models, as is often the case with individual hospitals or financial institutions, collaborating with diverse entities can enhance the efficacy of deep learning. Nevertheless, such extensive data aggregation poses significant privacy risks, as highlighted by Ren et al. \cite{ren2018querying}. This challenge has spurred advancements in the field of Federated Learning (FL) \cite{mcmahan2017communication}. FL enables participants to cooperatively train the model by synchronizing only locally trained model parameters without exposing the raw data. 
In a typical FL system, a central parameter server is employed to coordinate a large federation of participating nodes and collects gradients and model parameters from data nodes instead of gathering data to update the global model. However, the centralized FL paradigm can introduce security vulnerabilities and the peril of a single point of failure. As a countermeasure, the concept of a decentralized FL framework has been advanced. This alternative approach eliminates the central parameter server, instead implementing model aggregation across nodes in a distributed fashion, thereby circumventing the aforementioned risks inherent to the centralized structure. 

Gossip-SGD is a notable algorithm in DFL\cite{blot2016gossip, jin2016scale, ormandi2013gossip}. Nodes asynchronously update models and exchange with adjacent nodes, generating an exhaustive model for the network. However, when individual nodes only have access to non-IID data, making it difficult to reach a consensus on a global model due to varying gradients.

The challenge of handling non-IID data is a critical issue within the field of FL \cite{ZHU2021371}. Concurrently, several researchers are grappling with this problem, specifically within the paradigm of Decentralized FL. PDMM-SGD algorithm, delineated by Niwa et al.~(2020)\cite{niwa2020edge}, provides an effective method for training a robust global model under linear consensus constraints, which advocate for the uniformity of model parameters across all nodes. Nevertheless, in strongly non-IID settings, especially when the data bias is markedly high, this algorithm could face considerable difficulties in achieving a high-precision global model. 
Additionally, communication costs tend to be substantial due to frequent communication within the network. 

In response to the aforementioned challenges, we introduce a novel FL framework, designated as Tram-FL. 
Contrary to existing methodologies, Tram-FL facilitates learning by circulating a solitary global model amongst nodes. In other words, it foregoes the use of local models and model aggregation, with each node directly updating the global model instead. As a result, an appropriate selection of nodes for model traversal can emulate a state of training under IID data. For instance, by considering the class of data each node possesses and opting for nodes in a manner that approximates IID for the training data, Tram-FL can ameliorate accuracy degradation, even in exceedingly non-IID scenarios. Furthermore, given its faster model convergence compared to conventional DFL and the fact that it only requires a single model to be transmitted within the network, Tram-FL can facilitate the learning process with significantly reduced communication overhead. 

In Tram-FL, the determination of these routes for model traversal exerts a significant impact on both the model's convergence and the requisite communication traffic. This task of identifying the most suitable route is termed as the model routing problem. In order to resolve this problem, we propose a dynamic model routing algorithm that aims to achieve a highly precise global model with minimized model transmissions. Our proposed algorithm is predicated on the hypothesis that enhanced accuracy can be achieved with a reduced number of communication rounds by utilizing a training strategy that minimizes bias in the selection of both short-term and long-term samples. In selecting the subsequent node, the algorithm is designed to reduce the distribution bias inherent in the data hitherto employed for model training.

We conducted two experiments to evaluate the efficacy of Tram-FL. In the first experiment, we assessed the effectiveness of Tram-FL by comparing the accuracy achieved per total number of model transmissions with existing methods in different DFL scenarios. The results indicated that Tram-FL demonstrated stable convergence with fewer model transmissions in almost all scenarios. In our second experiment, we assessed the efficacy of our proposed dynamic model routing algorithm by juxtaposing the total count of model transmissions needed to attain a specified level of accuracy across a range of model routing algorithms. The results demonstrated that our proposed dynamic model routing algorithm reached the designated level of accuracy with the least aggregate number of model transmissions, thereby substantiating its efficiency.
\section{PROPOSED METHOD}
\subsection{System model}
This section presents the essential requirements for implementing the proposed method. In this study, we operate under the assumption of cross-silo FL. Unlike cross-device FL, which involves tens of thousands of devices such as smartphones, cross-silo FL pertains to a FL scenario where dozens of data servers participate \cite{kholod2020open}. Cross-silo FL contemplates a context where although data cannot be shared among nodes, such as in factories, hospitals, or banks, these nodes are mutually trustworthy. This setup facilitates cooperative training of deep learning models for applications such as pathological detection or fraudulent transaction detection. In this paper, we assume that the data servers involved in cross-silo FL, subsequently referred to as nodes, collectively train a deep neural network (DNN) model for a specific shared application.

It is assumed that the nodes possess suitably preprocessed data for their shared deep learning tasks. Specifically, each node is presumed to hold data that has been subjected to suitable preprocessing steps, such as labeling and completion of missing values, enabling it to train models using its own data. Nodes construct mini-batches from their data via random sampling for model training. This assumption is ubiquitous in the field of FL.

Moreover, it is anticipated that statistical information regarding the dataset owned by nodes, such as label distributions, can be pre-shared. This information is crucial for determining the root of the model in the proposed method, and it is an assumption often used in the FL setup under a non-IID setting \cite{Duan2019}.

Furthermore, this research assumes that the corporate data servers in cross-silo FL, denoted as nodes, are logically interconnected via the Internet, forming a fully meshed network topology. It is presumed that the logical links between nodes are stable, and each node can maintain uninterrupted and error-free communication with all other nodes at a satisfactory speed, perpetually.

\begin{figure}[t]
\begin{algorithm}[H]
    \caption{Tram-FL}
    \label{alg1}
    \begin{algorithmic}[1]
    \STATE Initialization of $w^{(0)}$
    \STATE Locate $w^{(0)}$ to randomly selected node $i$
    \FOR{each round $k = 1, 2, 3,\dots, K$}
        \STATE $\triangleright$ Step 1: Update model parameters
        \STATE $w^{k} \leftarrow w^{k-1} - \eta \nabla F(w^{k-1})$
        \IF{$k \mod T = 0$}
            \STATE $\triangleright$ Step 2: Transmit model to next node
            \STATE Select node $j \in \mathcal{V}$ by formula (5)
            %\STATE $\text{Transmit} _{i \rightarrow j}(w^{k+1})$
            \STATE Transmit $w^{k+1}$ to node $j$
        \ENDIF
    \ENDFOR
    \end{algorithmic}
\end{algorithm}
\end{figure}

\begin{figure}[!t]
\centering
\includegraphics[width=1\columnwidth]{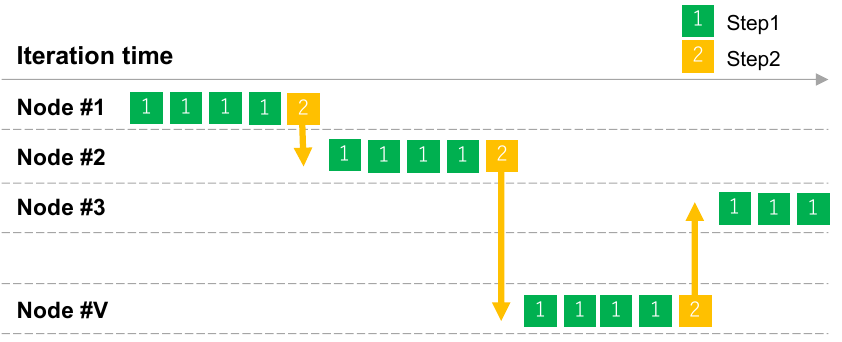}
\caption{Illustration of the sequence in the proposed Tram-FL model updating procedure. Step 1 involves the node updating the model parameter using a minibatch, followed by the transmission of the model to the next node in Step 2.}
\label{fig1}
\end{figure}

\subsection{Tram-FL Algorithm}
Tram-FL aims to achieve model convergence in non-IID settings with minimal communication costs where data imbalance among nodes is significantly high. To accomplish this objective, it is an important idea to circulate a single model throughout the entire network via an appropriate route during distributed model training.
    
\textbf{Fig.~1} delineates an exemplar of the sequence involving model updates and the subsequent model transfers amongst the nodes. A node that obtains a global model proceeds to update this model using a specific number of its data batches, as depicted in Step 1 of the figure. The quantity of batches deployed for these local updates is denoted as $T(\geq 1)$, exemplified by $T = 4$ in the figure. Subsequently, the global model is transferred to the next node, determined by the model routing algorithm, which is represented as Step 2 in the figure. This procedure is iteratively executed until the model's validation loss or accuracy reaches a predetermined threshold, or until a specified number of communication rounds are completed.
    
\textbf{Algolithm~1} presents the Tram-FL algorithm. The DNN model parameters at iteration $k$ are represented by $w^k$, and the loss function that requires minimization is denoted by $F(\cdot)$. Let $V$ denote the set of nodes, and $\eta$ be the learning rate. In each of the $K$ rounds of iteration, Step 1 and Step 2 are repeated. In Step 1, the node holding the model updates $w^k$ by extracting one batch of its own training data. This procedure is replicated $T$ times prior to advancing to Step 2, where the subsequent node designated for learning is determined and the model is accordingly transmitted.
The selection of the subsequent node is predicated on a model routing algorithm elaborated in Section~\ref{sec:routing}.

Our proposed framework curtails overall network traffic by transmitting a single model, which is advantageous compared to conventional approaches transmitting multiple models. However, a drawback inherent in this approach is the serialization of model updates, which can increase communication rounds and extend learning time.
Interestingly, the serialization of model updates also confers a substantial benefit in dealing with non-IID data, offsetting the increase in communication rounds. Even though each mini-batch may be non-IID during the training phase, our method—through consistent updating of the model with a selected subset of mini-batches and their distribution to a diverse set of nodes—approximates the performance of a model trained with IID data over medium-to-long-term periods. Notably, when both the batch size and the T parameter are set to 1, and with an appropriate selection of nodes, our methodology is congruent with training conducted using stochastic gradient descent on aggregated, or IID, data.

However, adjusting the batch size and T to 1 increases the frequency of model transmissions and network traffic. Additionally, biased node selection within the model's routing path could introduce considerable deviation from the ideal update scenario, or updates using IID data, potentially compromising the model's convergence capability. Therefore, optimizing batch size, T parameter, and the model's routing pathway is critical, but this study mainly focuses on enhancing the optimization of the model's routing, leaving batch size and T adjustment for future research. 

\subsection{Model Routing Algorithm}\label{sec:routing}
In this paper, we propose a dynamic model routing algorithm that decides the model's circulation path based on the label distribution of the data previously used for model updates and the label distribution of the data owned by the nodes. The algorithm is premised on the hypothesis that maximum accuracy can be accomplished with fewer communication rounds by conducting training that avoids the usage of biased samples in both short and long terms. Therefore, the proposed algorithm aims to minimize the bias in the label distribution of the data used for training. That is to say, it seeks to approximate a uniform distribution of labels by preferentially selecting nodes that possess data with underrepresented labels.

We define a round as a period during which a node receives a model, updates it, and transfers it to the subsequent node. Let's denote $L_i = \{l_{i,c} \mid c \in C\}$ as the number of training samples stored by node $i$, where $C$ signifies a set of labels and $l_{i, c} \geq 0$ indicates the quantity of samples with label $c$ present at node $i$. The total number of samples possessed by node $i$ is represented by $N_i$, hence $N_i = \sum_{c \in C} l_{i, c}$. Moreover, let $L^k_i$ represent the number of samples utilized by node $i$ for training the received model during round $k$, thus $L^k_i = \{l^k_{i,c} \mid c \in C\}$, with $l^k_{i,c} ,(0 \leq l^k_{i,c} \leq l_{i,c})$ being the count of samples with label $c$ used in round $k$.

We posit that the cumulative number of samples used for model training up to round $k$ is expressed as $L^k$, yielding
\begin{eqnarray}
L^k &=& \{\sum^{K}_{k=1}l^k{i_k, c} \mid c \in C\},
\end{eqnarray}
where $K$ represents the current round, and $i_k$ symbolizes the node possessing the model at round $k$.For simplified notation, we use $l^k_c$ to represent $\sum^{K}_{k=1}l^k{i_k, c}$.

Our proposed routing algorithm selects the succeeding node such that $L^k$ adheres to a pattern closely approximating a uniform distribution, specifically aiming to achieve 
\begin{equation}\label{eq:obj}
l^k_c = l^k_{c'} \; (c, c' \in C).
\end{equation}
To accomplish this, the proposed algorithm identifies the optimal node, $i*$, that is expected to minimize the variance of $L^k$ as the succeeding node. The selection policy can be formulated as follows:
\begin{eqnarray}\label{eq:opt}
i* &=& \underset{j}{\mathrm{argmin}}  \; \sigma (L^k + \frac{BT}{N_{j}} L_{j})\\
\sigma(L^k) &=& \frac{1}{|C|} \sum_{c \in C} (l^k_c - \frac{1}{|C|} \sum_{c \in C} {l^k_c}),
%\bar{l^k_c} &=& \frac{1}{|C|} \sum_{c \in C} {l^k_c},
\end{eqnarray}
where $B$ denotes the batch size. In this study, we assume that the node receiving the model randomly samples its (non-IID) data to construct mini-batches of size $B$, which are then used for training in that round. Consequently, $\frac{BT}{N_{j}} L_{j}$ reflects the expected number of samples labeled $c$ that will be utilized by node $j$ for training the model. Therefore, by determining the node according to \eqref{eq:opt}, the expected value of the variance of $L^{k+1}$ can be minimized. When the variance becomes 0, our objective \eqref{eq:obj} is fulfilled. That is, the distribution of $L^{k+1}$ will be a uniform distribution, which suggests that all labels have been used equally for model training up to round $k+1$. This uniform distribution is desirable as it implies unbiased learning based on all available labels, leading to a model with balanced performance across all classes.

\section{EXPERIMENTAL SETUP}
We conducted experimental evaluations across two scenarios, both operating under a non-IID setting. For the initial scenario, we benchmarked Tram-FL (without the model routing algorithm) against two baseline methods, Gossip-SGD \cite{blot2016gossip, jin2016scale, ormandi2013gossip} and PDMM-SGD \cite{niwa2020edge}. This comparison served to underscore the potential of Tram-FL to enhance model convergence, particularly in the context of non-IID data distributions.

\begin{table}
  \caption{Experimental setup}
  \label{tb: Experimental setup}
  \centering
  \begin{tabular}{l|lll}
  \hline
    Dataset & MNIST & CIFAR-10 & IMDb \\ \hline
    Number of nodes & 3, 5, 10 & 3, 5, 10 & 3, 5 \\ \hline
    Model size & 1,199,882 & 2,168,746 & 328,386 \\ \hline
    Input size & 784 & 1024 & 80 \\ \hline
    Mini-batch size & 100 & 100 & 100 \\ \hline
    Learning-rate & 0.005 & 0.005 & 0.05 \\ \hline
    Model transmission interval & 6 & 1 & 1 \\ \hline
    Total iteration number & 5,000 & 100,000 & 25,000 \\ \hline
  \end{tabular}
\end{table}

We employed two image classification tasks, namely MNIST\cite{lecun1998gradient} and CIFAR-10\cite{krizhevsky2010cifar}, and a text classification task called IMDb\cite{maas2011learning}. The MNIST dataset contains ten classes of handwritten digits, while the CIFAR-10 dataset comprises color images of objects belonging to ten different classes. The IMDb dataset consists of movie reviews that are labeled either positive or negative. The sample sizes of the training and test sets are {60,000, 10,000}, {50,000, 10,000}, and {25,000, 25,000} for MNIST, CIFAR-10, and IMDb, respectively. The training samples were distributed to the nodes in the following way.
\vspace{2mm}

\noindent\textbf{MNIST, CIFAR-10}: The data labels assigned to each node are distributed uniformly and without any overlapping. Specifically, when $V=3$, the labels are partitioned into three sets $\{0,1,2\}$, $\{3,4,5\}$, and $\{6,7,8,9\}$, and assigned to node 0, 1, and 2, respectively. In a similar fashion, when $V=5$, the labels are partitioned into five sets $\{0,1\}$, $\{2,3\}$, $\{4,5\}$, $\{6,7\}$, and $\{8,9\}$, while for $V=10$, they are partitioned into ten sets $\{0\}$, $\{1\}$, $\{2\}$, $\{3\}$, $\{4\}$, $\{5\}$, $\{6\}$, $\{7\}$, $\{8\}$, and $\{9\}$.
Each node is assigned all the samples that have data labels distributed in that node. In MNIST, when $V=3$, $N_1=18000$, $N_2=18000$, $N_3=24000$. When $V=5$, $N=12000$ and when $V=10$, $N=6000$. In CIFAR-10, when $V=3$, $N_1=15000$, $N2=15000$, $N3=20000$. When $V=5$, $N=10000$ and when $V=10$, $N=5000$.
\vspace{2mm}

\noindent\textbf{IMDb}: The training data is divided based on the exponential distribution and its corresponding cumulative distribution function. Specifically, when V=3, $L_0$, $L_1$, and $L_2$ were [10,125, 2,625], [2,000, 4,750], and [375, 5,125], respectively. Similarly, when V=5, $L_0$, $L_1$, $L_2$, $L_3$, and $L_4$ were [7,875, 1,125], [2,875, 2,375], [1,125, 2,875], [375, 3,000], and [250, 3,125], respectively.\\

A DNN architecture for each task was as follows: The model for MNIST was a convolutional neural network (CNN) comprising two 3 × 3 convolutional layers with 32 and 64 output channels, respectively, which are activated by Rectified Linear Units (ReLU). Following the convolutional layers are 2 × 2 max pooling and a dropout rate of 0.5. Subsequently, the architecture incorporates two fully connected layers with 128 units activated by ReLU and 10 units with a dropout rate of 0.5 in between. The model for CIFAR-10 consists of four 3×3 convolutional layers with 32, 64, 64, and 64 channels, each activated with ReLU and group normalized. Further, the model includes 2×2 per 2-layer max pooling and a dropout rate of 0.25. Additionally, the architecture has two all-coupled layers with 512 units activated with ReLU and 10 units activated with softmax, with a dropout rate of 0.5 in between. For the IMDb, we used LSTM, a type of recurrent neural network (RNN) known for its ability to handle sequential data. Specifically, we used the same model architecture as the one used in the Keras tutorial\cite{chollet2015keras} for the IMDb dataset. The model includes an embedding layer with 32 output dimensions for each word, an LSTM layer with 32 nodes, and a fully connected layer with two units activated with softmax. He's method\cite{he2015delving} was utilized to initialize w with a shared random seed for each node.

We set the number of nodes to range from 3 to 10. We assumed full-mesh connectivity, wherein all nodes are adjacent to one another. Therefore, in both Gossip SGD and PDMM-SGD, each round involves an exchange of the model with all neighboring nodes. In contrast, for our proposed method, we structured the model to initiate from node 0 and then transmit sequentially to nodes 1, 2, 3, etc. After reaching the last node, the model is looped back to node 0, and this cycle is repeatedly performed.
The other hyperparameters are summarized in \textbf{TABLE~I}.

In the second scenario, we evaluated the effectiveness of the proposed model routing algorithm. We used the CIFAR-10 dataset while keeping all other settings the same except for data distribution. A mesh network with 5 nodes was emulated, and the data labels were randomly assigned to each node such that each node had between 2-5 labels. The samples were distributed to each node according to the label assignment. Each node is assigned all the samples that have data labels distributed in that node. Therefore, the number of samples increases in proportion to the number of assigned labels. We evaluated the convergence speed of the proposed dynamic routing and static routes for circulating models by assessing the number of required model transmissions to achieve a certain accuracy threshold. \textbf{TABLE~I\hspace{-.1em}I} presents the static routes. Routes 1 to 24 correspond to static model routing, constituting a total of 24 distinct routes that traverse all five nodes. We also used random routing as a baseline for the dynamic routing algorithm, in which the next node is selected randomly from among its neighbors.

\begin{table}[t]
  \caption{Comparison route}
  \label{Comparison route}
  \centering
  \begin{tabular}{ll|ll|ll}
  \hline
    Route~1 & 0,1,2,3,4 & Route~9 & 0,2,3,1,4 & Route~17 & 0,3,4,1,2 \\
    \hline
    Route~2 & 0,1,2,4,3 & Route~10 & 0,2,3,4,1 & Route~18 & 0,3,4,2,1 \\ \hline
    Route~3 & 0,1,3,2,4 & Route~11 & 0,2,4,1,3 & Route~19 & 0,4,1,2,3 \\ \hline
    Route~4 & 0,1,3,4,2 & Route~12 & 0,2,4,3,1 & Route~20 & 0,4,1,3,2 \\ \hline
    Route~5 & 0,1,4,2,3 & Route~13 & 0,3,1,2,4 & Route~21 & 0,4,2,1,3 \\ \hline
    Route~6 & 0,1,4,3,2 & Route~14 & 0,3,1,4,2 & Route~22 & 0,4,2,3,1 \\ \hline
    Route~7 & 0,2,1,3,4 & Route~15 & 0,3,2,1,4 & Route~23 & 0,4,3,1,2 \\ \hline
    Route~8 & 0,2,1,4,3 & Route~16 & 0,3,2,4,1 & Route~24 & 0,4,3,2,1 \\ 
    \hline
  \end{tabular}
\end{table}
\begin{comment}
この章では２つのシナリオでの評価実験の結果を示す。

認識精度およびクロスエントロピーのコストは、テストデータセットを用いて評価スコアを算出した。評価方法は「モデルの更新回数と通信回数」に対する認識精度とした。
また通信コストの削減に関して評価するためにネットワーク全体での総通信回数に対する認識精度を算出した。
図２に「更新回数と通信回数」に対する認識精度の評価を示す。
MNIST及びCIFAR-10の実験結果を図2(a)(b)に示す。提案手法はすべての設定で安定した収束性を実現した。図２(c)にIMDbの実験結果を示す。PDMM-SGDとGossip-SGDは少なくともこの実験における更新回数では収束しなかったが提案手法は収束した。
図３にネットワーク全体での総通信回数に対する認識精度の評価を示す。MNIST及びCIFAR-10、IMDbの実験結果を図3(a)(b)(c)に示す。提案手法はすべての設定において非常に少ない通信回数での収束を実現した。

図2は、ネットワーク内のモデル送信総数に対するテスト精度を示しています。ノード間のモデル送信総数が増えると、モデルが更新されて学習が進み、その結果、すべてのタスクでテスト精度が向上していることがわかります。
MNISTとCIFAR-10の実験結果をそれぞれ図2(a),(b)に示す。ノード数やデータ分布のばらつきにより収束が不安定になり、グローバルモデルを達成できない比較手法もある中、Tram-FLはすべての設定でグローバルモデルを達成することができた。一方、図2(c)はIMDbの実験結果であり、PDMM-SGDとGossip-SGDはこの実験の送信回数内で収束しなかったが、提案手法は収束させた。

評価方法はモデルの認識精度が78％に達した時点での更新回数とした。評価結果を図３に示す。提案したmodel routing algorithmは最も少ない更新回数で精度78％に達した。また、static model routing algorithmと比較して平均で16.25%少ない更新回数となった。dynamic model routing algorithmとして使用した巡回経路を一様ランダムに選択するものと比較しても平均で20.22%少ない更新回数を達成した。
表３はテスト精度78%に達するまでのモデル送信回数の平均と標準偏差を示している。Static routingは最も良い経路と中央値のルート、最も悪いルートの３つに関して調べた。提案したmodel routing algolithmが最も少ないtotal number of model transmissionsで精度７８％を達成した。
提案したmodel routing algolithmによってglobal modelに到達するまでの通信コストを削減することができた。

\end{comment}

\section{Experimental Result}
\begin{comment}
An evaluation score was calculated using the test data set to assess recognition accuracy and the cost of cross-entropy. The evaluation method was based on recognition accuracy with respect to the number of model updates and communications. Furthermore, the recognition accuracy was calculated for the total number of communications in the entire network to evaluate the reduction of communication costs.
\textbf{Fig. 2} presents the evaluation results of recognition accuracy for the number of updates and communications. Experimental results for MNIST and CIFAR-10 are shown in \textbf{Fig. 2(a),(b)}, respectively, demonstrating stable convergence in all settings. In contrast, \textbf{Fig. 2(c)} shows the experimental results for IMDb, where PDMM-SGD and Gossip-SGD failed to converge within the number of updates in this experiment, while the proposed method achieved convergence.
\textbf{Fig. 3} illustrates the evaluation of recognition accuracy versus the total number of communications in the entire network. The experimental results for MNIST, CIFAR-10, and IMDb are presented in \textbf{Fig. 3(a),(b),(c)}, respectively. The proposed method achieved convergence with a minimal number of communications in all settings.
\end{comment}
\noindent \textbf{Scenario 1:}
\textbf{Fig. 2} shows the test accuracy against the total number of model transmissions in the network for the first scenario. As the total number of model transmissions between nodes increases, the models are updated and learning progresses, and as a result, the test accuracy increases in all the tasks. 
Experimental results for MNIST and CIFAR-10 are shown in \textbf{Fig. 2(a),(b)}. Despite some comparative approaches being unable to obtain an accurate global model owing to unstable convergence resulting from variations in the number of nodes and data distribution, Tram-FL managed to obtain an accurate global model in all settings. In contrast, \textbf{Fig. 2(c)} shows the experimental results for IMDb, where PDMM-SGD and Gossip-SGD failed to converge within the number of transmissions in this experiment, while the proposed method obtained an accurate global model.

\setlength\intextsep{0pt}
\begin{figure}[!t]
    \centering
    \begin{tabular}{cc}
    \subfloat[MNIST]{\includegraphics[width=0.48\columnwidth]{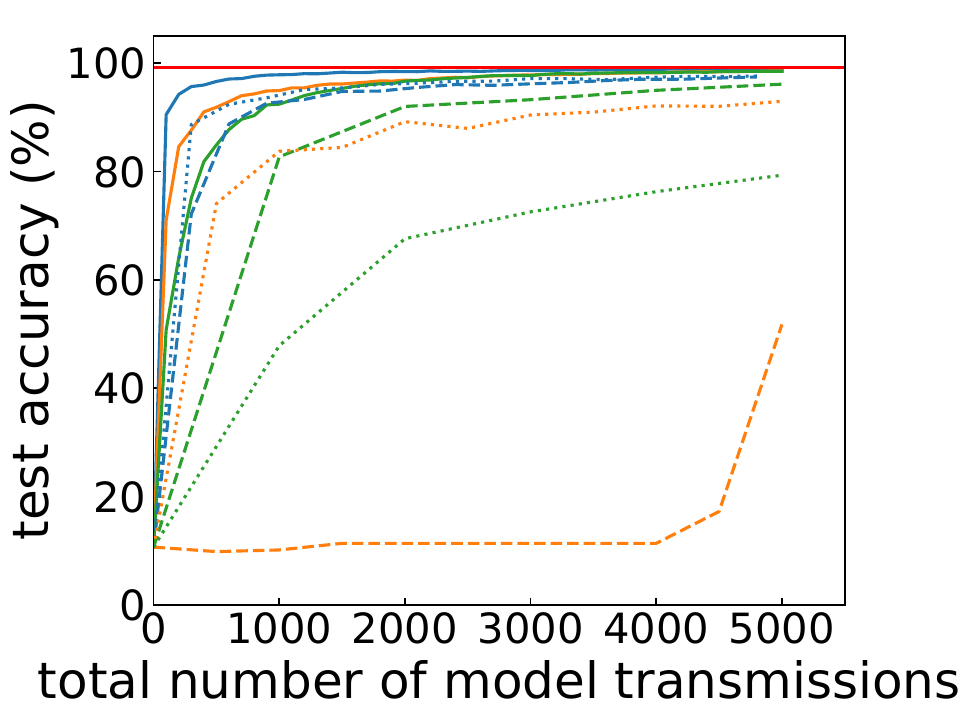}} &
    \subfloat[CIFAR-10]{\includegraphics[width=0.48\columnwidth]{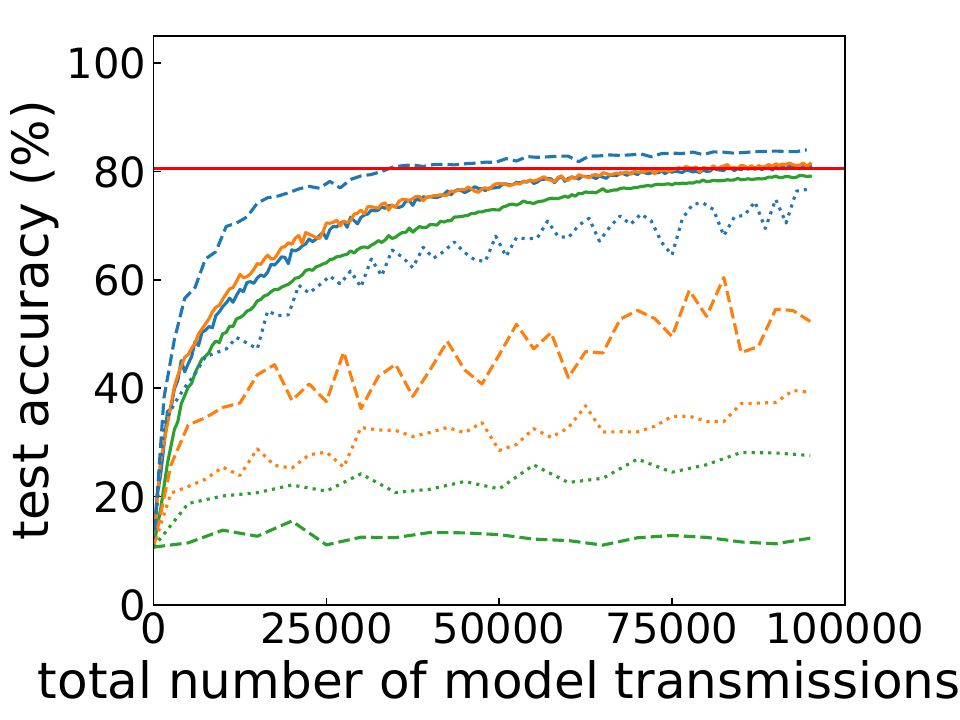}} \\
    \centering
    \subfloat[IMDb]{\includegraphics[width=0.48\columnwidth]{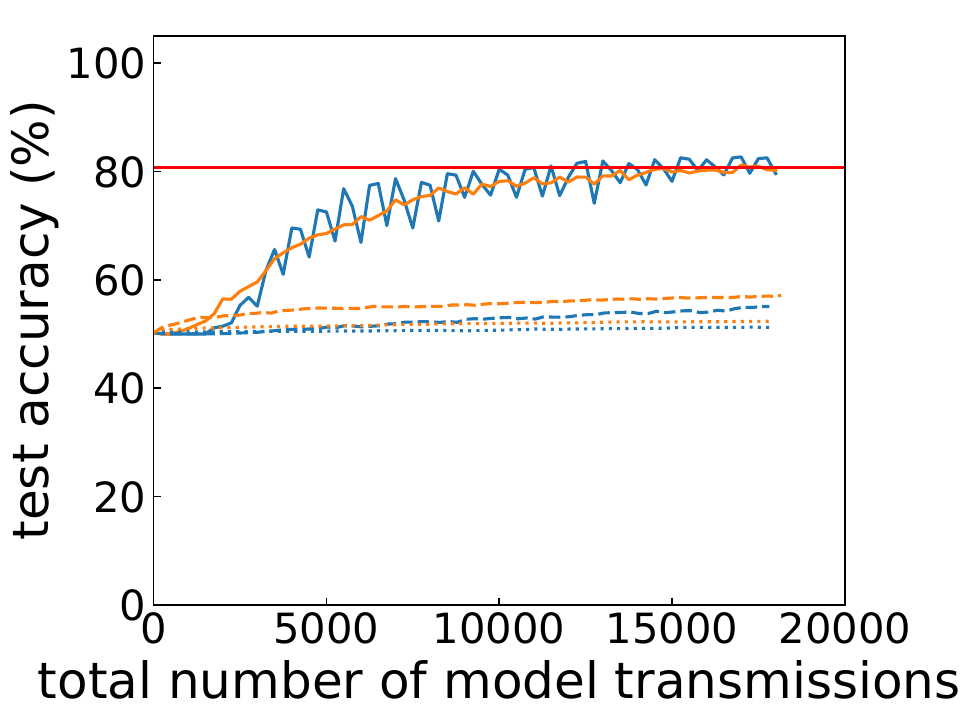}} &
    \subfloat{\includegraphics[width=0.28\columnwidth]{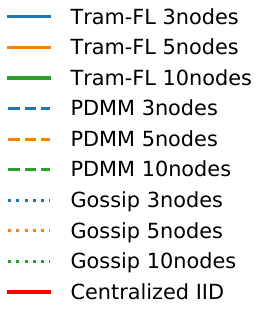}}
    \end{tabular}
    \caption{Test accuracy as a function of the total number of model transmissions, utilizing non-IID data sets. Evaluation includes (a) MNIST for handwritten digit classification, (b) CIFAR-10 for object recognition, and (c) IMDb for sentiment analysis.}
    \label{fig3}
\end{figure}

\begin{figure}[!t]
  \centering
  \includegraphics[width=1\columnwidth]{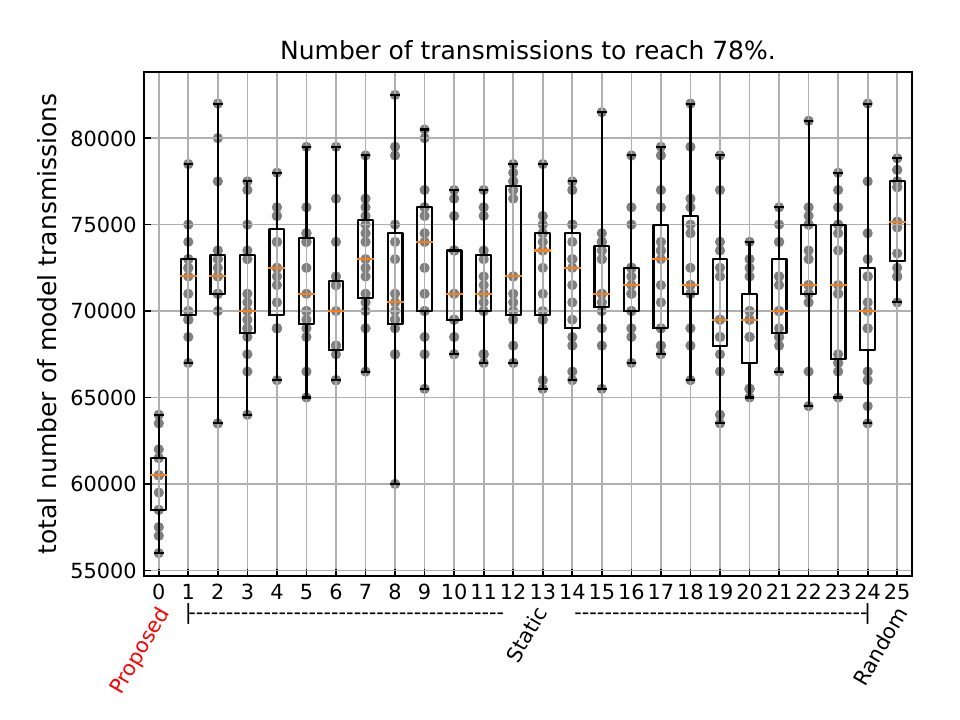}
  \caption{Comparison of the number of model transmissions required to reach a 78\% accuracy level, illustrated across different model routing methods.}
  \label{fig4}
\end{figure}

\vspace{2mm}
\noindent \textbf{Scenario 2:}
\textbf{Fig. 3} presents a box plot displaying the total number of model transmissions required to attain a test accuracy of 78\% -- a mark 2\% lower than the one achieved by centralized learning for CIFAR-10 -- in Tram-FL, using each routing method. This data is drawn from a specific distribution and is based on 15 trials.
Compared to static model routing, the proposed routing algorithm demonstrated an average reduction of 16.25\% in model transmissions. Furthermore, when compared to the random routing, the proposed algorithm exhibited an average decrease of 20.22\% in model transmissions. 
\textbf{TABLE~I\hspace{-.1em}I\hspace{-.1em}I} presents the mean and standard deviation of the total number of model transmissions required to achieve a test accuracy of 78\%. We evaluated three routes for static routing: the best route, median route, and worst route, which were identified via a comprehensive search. Our proposed model routing algorithm achieved a test accuracy of 78\% with the least total number of model transmissions. The efficiency realized through our proposed method can be attributed to its dynamic routing nature. While static routing entails the selection of each node in an orderly, evenly spaced manner, our method permits the consecutive or biased selection of the same node. This flexibility allows for a closer approximation of a uniform data distribution, a strategic approach absent in static routing schemes.

\begin{table}[!t]
  \caption{Average number of model transmissions}
  \label{tb: Average number of model transmissions pm std.}
  \centering
  \begin{tabular}{lccccc}
  \toprule
    Proposed & \multicolumn{3}{c}{Static routing} & Uniform random \\
    \cmidrule(lr){2-4}
    & Best & Median & Worst & \\ \midrule
    60778 & 64667 & 72722\ & 80444 & 71611 \\
    $\pm$3119 & $\pm$4435 & $\pm5$135 & $\pm$5535 & $\pm$6293 \\ \bottomrule
  \end{tabular}
\end{table}

\begin{comment}

本稿では, モデルの巡回に基づく新たなP2P FL手法(Tram-FL)を提案し、dynamic model routing algorithmの一例を示した。

Tram-FLはネットワーク全体で1つのみモデルを共有する。このモデルを適切な経路で巡回させることで学習を進め、グローバルモデルを実現する。適切な巡回経路は事前に共有される各クライアントのデータラベル分布をもとにmodel routing algorithmによって決定される。

提案アルゴリズムは非IIDデータセットをノードに配置しても、さまざまなシナリオにおいて最小限の通信コストで安定した収束性を実現した。また提案したmodel routing algorithmはstatic model routing algorithmや一様ランダムな経路と比較して、少ない通信回数でモデルを収束させた。これは、最適な経路を選択することができるmodel routing algorithmを確立すれば非IIDデータセットに対して無駄のない学習が可能なことを示唆する。提案したmodel routing algorithmよりも早く収束するmodel routing algorithmを確立することは今後の課題である。
さらに、提案するdynamic model routing algolithmでは学習に使用されるラベルにのみ着目するため、学習に参加するノードの偏りが大きくなることがある。全てのノードが平等に学習に参加するようなalgorithmを考えることは興味深い今後の研究である。
\end{comment}

\section{Conclusion}
In this paper, we introduced a novel decentralized FL approach that utilizes model traversal, dubbed Tram-FL. We also present an example of a dynamic model routing algorithm. Our proposed algorithm obtains an accurate global model with minimal communication costs in diverse scenarios, even when non-IID datasets are distributed among nodes. Notably, the proposed model routing algorithm exhibits faster convergence and requires fewer transmissions than both the static model routing algorithm and uniform random routes. This indicates that developing a model routing algorithm that selects the optimal route can facilitate efficient learning on non-IID datasets. Our future research plans involve the development of an enhanced routing algorithm that optimizes both the batch size and the number of mini-batches. This enhancement is expected to further reduce the number of required communication rounds, thereby improving overall network efficiency and model performance.

\bibliographystyle{IEEEtran}
\bibliography{main}

\end{document}